\pdfoutput=1
\documentclass{article}

\PassOptionsToPackage{numbers, compress}{natbib}
\usepackage[preprint]{neurips_2026}
\usepackage{comment}
\usepackage{algorithm}
\usepackage{algpseudocode}

\usepackage[utf8]{inputenc} 
\usepackage[T1]{fontenc}    
\usepackage{hyperref}       
\usepackage{url}            
\usepackage{booktabs}       
\usepackage{amsfonts}       
\usepackage{nicefrac}       
\usepackage{amsmath}
\usepackage{microtype} 
\usepackage{algorithm}
\usepackage{algpseudocode} 
\usepackage{xcolor}         
\usepackage{graphicx}
\title{Neural-Schwarz Tiling for Geometry-Universal PDE Solving at Scale}

\author{%
  Paolo Secchi\textsuperscript{1} \quad
  Daniel S. Balint\textsuperscript{1} \quad
  Marco Maurizi\textsuperscript{2} \\[0.5em]
  \textsuperscript{1}Imperial College London \\
  \textsuperscript{2}Italian Institute of Artificial Intelligence \\
  \texttt{p.secchi22@imperial.ac.uk}\\
  \texttt{marco.maurizi@ai4i.it}
}

\begin{document}

\maketitle

\begin{abstract}
Most learned PDE solvers follow a global-surrogate paradigm: a neural operator is trained to map full problem descriptions to full solution fields for a prescribed distribution of geometries, boundary conditions, and coefficients. This has enabled fast inference within fixed problem families, but limits reuse across new domains and makes large-scale deployment dependent on expensive problem-specific data generation. We introduce $\textbf{NEST}$ ($\textbf{Ne}$ural-$\textbf{S}$chwarz $\textbf{T}$iling), a local-to-global framework that shifts learning from full-domain solution operators to reusable local physical solvers. The central premise is that, although global PDE solutions depend on geometry, scale, and boundary conditions, the physical response on small neighborhoods can be learned locally and composed into global solutions through classical domain decomposition. NEST learns a neural operator on minimal voxel patches ($3 \times 3 \times 3$) with diverse local geometries and boundary/interface data. At inference time, an unseen voxelized domain is tiled into overlapping patches, the learned local solver is applied patchwise, and global consistency is enforced through iterative Schwarz coupling with partition-of-unity assembly. In this way, generalization is shifted from a monolithic neural model to the combination of local physics learning and algorithmic global assembly. We instantiate NEST on nonlinear static equilibrium in compressible neo-Hookean solids and evaluate it on large, geometrically complex 3D domains far outside the scale of the training patches. Our results show that local neural building blocks, coupled through Schwarz iteration, offer a reusable local-training path toward scalable learned PDE solvers that generalize across domain size, shape, and boundary-condition configurations.
\end{abstract}

\section{Introduction}
\label{sec:intro}
Machine learning has transformed domains in which data admit a universal representation. 
In language, large-scale models trained on token sequences generalize across a broad range of tasks, domains, and contexts \cite{brown2020language}. 
This success rests on a common compositional substrate: sentences, documents, code, and instructions can all be represented as sequences of tokens, enabling a shared modeling paradigm across heterogeneous problems.

No comparable foundation has emerged for physical simulation. 
Many important problems in science and engineering---fluid flow, heat transfer, elasticity, fracture, electromagnetism, and coupled multiphysics---are governed by partial differential equations (PDEs). 
These describe fields evolving over geometries, boundary conditions, material distributions, and forcing terms that vary dramatically from one problem to another. 
Unlike language, PDE problems are tied to their spatial domain, discretization, boundary representation, and physical coefficients. 
The central object of physical simulation is thus not a single task, but a family of operators whose structure changes with the geometry and physics of the problem.

This lack of universality is a major bottleneck for scientific machine learning (SciML) \cite{subramanian2023towards,mccabe2024multiple,tierz2025feasibility}. 
Classical numerical solvers such as finite element modeling (FEM) are general and reliable, but their cost grows rapidly with problem size, resolution, and the number of repeated solves required by downstream workflows like inverse design, uncertainty quantification, digital twins, or design-space exploration \cite{karniadakis2021physics,maurizi2025designing}. 
Learned surrogates can be fast after training, but are typically specialized: a model trained on one geometry, discretization, or coefficient distribution often fails to transfer outside that regime. 
Consequently, today's PDE learning pipelines replace one expensive computation with another: generating large datasets and retraining specialized models for each problem family. 
Physics-informed neural networks (PINNs) \cite{raissi2019physics} sidestep data generation by embedding the PDE residual into the loss, but each new geometry, boundary condition, or coefficient field requires re-optimizing the network from scratch, so generalization across problems is essentially absent.

The key challenge is therefore not to accelerate a fixed benchmark, but to learn reusable physical computation that generalizes across domains and scales with problem size \cite{ma2026learning,masliaev2025towards,zhu2025generalizing}. 
A general learned solver should not need retraining whenever the geometry, boundary conditions, or domain size change; it should instead learn local physical laws composable into global solutions on previously unseen structures. 
This is the central question of this work: can we learn a small, reusable neural building block for PDE solution, and assemble it algorithmically to solve large, geometrically arbitrary physical systems?

SciML \cite{ma2026learning} often formulates simulation as learning a parameterized surrogate operator
$\mathcal{G}_\theta \approx \mathcal{G}^{\dagger}: \mathcal{X} \rightarrow \mathcal{U}$,
where $\mathcal{G}^{\dagger}$ maps functional inputs --- such as boundary data or initial conditions --- to solution fields.
Neural operators \cite{kovachki2023neural} have emerged as a principled class of models for approximating infinite-dimensional mappings, typically of the form
$$
\mathcal{G}_\theta
=
\mathcal{Q} \circ \sigma\left(W_L+\mathcal{K}_L\right)
\circ \cdots \circ
\sigma\left(W_1+\mathcal{K}_1\right)
\circ \mathcal{P},
$$
where $\mathcal{P}, \mathcal{Q}$ are lifting and projection maps, $\mathcal{K}_{\ell}$ are learned integral kernel operators, $W_\ell$ are local linear maps, and $\sigma$ is a pointwise nonlinearity.
They act on functions rather than fixed finite-dimensional vectors and admit guarantees of discretization invariance and universal approximation \cite{kovachki2023neural}.

However, these guarantees apply to operators between function spaces on fixed domains. 
In practice, a learned operator is trained for a prescribed geometry class, boundary representation, and coefficient distribution. 
While several architectures improve empirical transfer across meshes, resolutions, and parameter fields, the learning problem remains global: the model approximates an entire PDE solution operator over a given problem family. 
Changing the domain, enlarging the computational region, or modifying boundary-condition structure typically requires new data generation, retraining, or architecture-specific adaptation, limiting neural operators as reusable computational primitives.

In this work, we take a different view.
Rather than learning a global operator for each class of PDE problems, we learn a local physical solver once and reuse it compositionally.
The core idea is to train a neural operator on small voxelized subdomains, exposing it to a rich distribution of boundary/interface conditions.
At test time, arbitrary geometries are voxelized into overlapping local neighborhoods, and global consistency is enforced through Schwarz iteration \cite{toselli2004domain,mathew2008domain}.
Generalization thus shifts from the neural model alone to the combination of a learned local solution operator and a classical domain-decomposition procedure.

\paragraph{Our main contributions.}
We introduce \textbf{NEST} (\textbf{Ne}ural-\textbf{S}chwarz \textbf{T}iling), a local-to-global framework for solving PDEs on previously unseen domains without retraining.
NEST learns a reusable neural building block on small voxel neighborhoods and assembles these blocks through overlapping Schwarz iteration to obtain global solutions on large, geometrically complex structures.
This construction separates local physics learning from global domain assembly: the neural operator approximates local solution updates, while the Schwarz procedure propagates information, enforces compatibility across overlapping subdomains, and enables scaling to domains much larger than those seen during training.

Our contributions are threefold.
First, we propose a neural-Schwarz formulation in which a single local operator, trained on small voxel patches with diverse boundary conditions, can be reused across arbitrary voxelized geometries.
Second, we show that global PDE solutions can be recovered by composing these learned local updates through Schwarz iteration, enabling inference on domains whose size, shape, and topology differ from the training setting.
Third, we demonstrate that this paradigm provides a reusable local-training alternative to training specialized global surrogates for each new geometry class or problem family, opening a path toward learned PDE solvers that behave less like task-specific predictors and more like reusable computational building blocks.

\section{Related Work}

\paragraph{Neural operators for parametric PDEs.}
Operator learning has emerged as a central paradigm for scientific machine learning, aiming to approximate mappings between infinite-dimensional function spaces rather than finite-dimensional input-output pairs.
Early architectures such as DeepONet \cite{lu2021learning} and the Fourier Neural Operator (FNO) \cite{li2020fourier} demonstrated that neural networks can learn solution operators for families of parametric PDEs, enabling fast inference after training.
Subsequent work developed a broader theory of neural operators, including universal approximation and discretization-invariance results \cite{kovachki2023neural}, as well as physics-informed variants that incorporate PDE residuals or physical constraints during training \cite{li2024physics}.
These methods have significantly expanded the scope of learned PDE solvers, but they typically learn a global operator associated with a prescribed problem family.
In particular, the training distribution usually fixes or strongly constrains the domain class, boundary representation, coefficient statistics, and input-output structure.
As a result, changing the geometry or boundary-condition structure often requires new data, retraining, or architecture-specific adaptation.

\paragraph{Learning PDEs on meshes and irregular geometries.}
A complementary line of work addresses the limitations of grid-based operator learning by representing physical systems on meshes, graphs, point clouds, or geometry-aware latent spaces.
MeshGraphNets \cite{pfaff2021learning} introduced a graph-network simulator that performs message passing on adaptive mesh representations and demonstrated strong performance on deformable solids, fluids, and cloth.
Graph-based and geometry-informed neural operators extend this idea to operator learning on irregular domains, using graph kernels, point-cloud representations, signed-distance functions, or learned mappings between irregular and regular domains \cite{li2023geometry}. More recently, transformer architectures have advanced PDE modeling; Transolver \cite{wu2024transolver,luo2025transolver++} handles complex geometries via physics-aware attention mechanisms \cite{vaswani2017attention}, whereas HAMLET \cite{bryutkin2024hamlet} tackles parametric problems using graph attention.

These approaches are important steps toward geometry-aware and parametric learned simulation. However, they still primarily follow a global learning paradigm: the model is trained to map from a full geometry and its associated physical inputs to a full solution field.
Thus, generalization to geometries, topologies, domain sizes, or boundary-condition configurations far outside the training distribution remains difficult.
Moreover, scaling to very large domains can require substantially larger models, more memory, or additional training data.
In contrast, our goal is not to learn a single global operator over an entire geometry distribution, but to learn a reusable local solver that can be composed algorithmically into global solutions.

\paragraph{Domain decomposition and Schwarz methods.}
Domain decomposition methods provide a classical route for solving PDEs on large or complex domains by decomposing the global problem into smaller subproblems.
Among these, overlapping Schwarz methods iteratively solve local problems on overlapping subdomains and exchange interface information until a globally consistent solution is obtained \cite{dolean2015introduction}.
These methods are attractive because they expose locality, parallelism, and modularity: global solution structure emerges from repeated local solves and communication across overlaps.
Recent work has begun to combine domain decomposition with machine learning, for example by learning interface conditions, accelerating local solvers \cite{ouyang2026noem}, or coupling neural operators with Schwarz-type iterations \cite{taghibakhshi2022learning,huang2025operator,wu2026learning}. Such hybrid approaches suggest that classical numerical algorithms can provide useful structure for learned PDE solvers, especially when extrapolation beyond a fixed training domain is required. However, existing studies focus primarily on 2D domains, and extending these frameworks to 3D problems is not straightforward. Current methods often rely on learning non-minimal subdomains, which makes it difficult to tile complex 3D geometries \cite{huang2025operator}. Furthermore, the reliance on these specific subdomain configurations is computationally intensive and difficult to tailor to diverse problem sets such as PDE parameters.

\paragraph{Positioning of NEST.}
NEST builds on this local-to-global perspective, but differs from existing neural operators and neural domain-decomposition approaches in its use of a small, reusable voxel-level building block.
Instead of training a model on full domains \cite{li2023fourier,li2023geometry,li2025geometric} or geometry-specific decompositions \cite{ouyang2026noem}, NEST trains a local neural operator on small voxel neighborhoods with diverse boundary data and interface conditions.
At inference time, arbitrary voxelized domains are tiled into overlapping neighborhoods, the same learned operator is applied repeatedly, and Schwarz iteration enforces global consistency.
This separates the learning problem from the global geometry: the neural model learns local physical response, while the Schwarz procedure performs global assembly.
As a result, NEST aims to provide a path toward learned PDE solvers that are reusable across domain size, shape, topology, and boundary-condition configurations, rather than specialized to a fixed benchmark or geometry class.

\section{Preliminaries and Problem Statement}
We consider a general boundary value problem on a bounded domain $\Omega \subset \mathbb{R}^d$ with boundary $\partial \Omega$: given a (possibly nonlinear) differential operator $\mathcal{L}_\phi$ parameterized by $\phi \in \Phi$ (e.g.\ material constants, coefficients) and a prescribed Dirichlet boundary condition $g: \partial \Omega \rightarrow \mathbb{R}^m$, find $u: \Omega \rightarrow \mathbb{R}^m$ such that
\begin{equation}
\mathcal{L}_\phi[u]=0 \text{ in } \Omega, \quad u=g \text{ on } \partial \Omega,
\label{eq:problem}
\end{equation}
where the Dirichlet boundary condition is imposed on the entire boundary $\partial \Omega$. Our goal is to learn a surrogate solution operator
\begin{equation}
\mathcal{S}_\theta:(\Omega, g) \mapsto u
\end{equation}
that generalizes across both $g$ and the geometry $\Omega \in \mathcal{D}$, where $\mathcal{D}$ is a family of admissible domains rather than a fixed reference shape. Crucially, training $\mathcal{S}_\theta$ must be cheap enough that the surrogate can be easily retrained for new operator parameters $\phi$, enabling rapid tailoring to different PDE problems.

To illustrate NEST, we instantiate~\eqref{eq:problem} on \emph{quasilinear elliptic boundary value problems in divergence form},
\begin{equation}
\mathcal{L}_\phi[u] = -\operatorname{div}\bigl(\sigma_\phi(\nabla u)\bigr) = 0 \text{ in } \Omega, \quad u = g \text{ on } \partial \Omega,
\label{eq:divform}
\end{equation}
where $\sigma_\phi : \mathbb{R}^{m \times d} \to \mathbb{R}^{m \times d}$ is a (generally nonlinear) constitutive map. Different choices of $\sigma_\phi$ recover a wide range of stationary physical models such as heat conduction with temperature-dependent conductivity, nonlinear electrostatics, Darcy flow in heterogeneous media, and finite-strain solid mechanics. In this work, as a representative benchmark, we consider the \emph{nonlinear static equilibrium of a compressible neo-Hookean solid}, with $d = 3$, $m = 3$, and $\phi = (\lambda, \mu)$ the Lamé constants. Let $u$ denote the displacement, $F = I + \nabla u$ the deformation gradient, and $J = \det F$. The strain energy density
\begin{equation}
\psi_\phi(F) = \frac{\mu}{2}\left(\operatorname{tr}\left(F^{\top} F\right) - 3\right) - \mu \log J + \frac{\lambda}{2}(\log J)^2
\label{eq:psi}
\end{equation}
yields the first Piola--Kirchhoff stress $P_\phi(F) = \partial \psi_\phi / \partial F$, and identifying $\sigma_\phi(\nabla u) = P_\phi(I + \nabla u)$ recovers the divergence-form structure of~\eqref{eq:divform}. In the absence of body forces and surface tractions, the strong form reduces to
\begin{equation}
\operatorname{div} P_\phi(I + \nabla u) = 0 \text{ in } \Omega, \quad u = g \text{ on } \partial \Omega,
\label{eq:strong}
\end{equation}
which is the instance of~\eqref{eq:problem} on which we benchmark NEST. Solutions are computed with the FEniCS finite-element library~\citep{logg2012automated}.

\section{NEST --- NEural-Schwarz Tiling}
\label{sec:nest}
NEST is a two-stage framework for solving PDEs on large, geometrically complex, arbitrary domains by combining a learned local neural operator with a global Schwarz coupling strategy. Instead of training a neural operator directly on full-domain solutions for a prescribed family of geometries and boundary conditions, NEST learns a reusable local surrogate, $\mathcal{S}_\theta$, on a small canonical patch. The local operator maps patch-level problem data---including geometry and boundary/interface conditions---to the corresponding local solution field.

At inference time, an arbitrary domain $\Omega \in \mathcal{D}$ is decomposed into overlapping patches. The learned operator $\mathcal{S}_\theta$ is applied independently on each patch, while neighboring predictions are coupled through an additive Schwarz iteration. Interface information is exchanged across overlapping regions, so that local solutions are progressively made consistent with one another and with the prescribed global boundary conditions. In this way, NEST turns a neural operator trained only on local building blocks into a solver that can be applied to domains and boundary conditions that are much larger and more complex than those seen during training. The following subsections describe the local neural operator, the tiling of arbitrary domains into overlapping patches, and the Schwarz-based global coupling procedure.

\subsection{Local Operator Learning}
\label{sec:local}
We restrict $\mathcal{S}_\theta$ to domains carved out of a fixed canonical patch $\Omega_0 = [0,1]^3$, which we discretize as a regular $3 \times 3 \times 3$ grid of hexahedral cells $\{C_k\}_{k=1}^{27}$. This is the smallest cubic voxel patch containing a single fully interior cell, making it the minimal hexahedral patch that supports overlapping Schwarz with non-trivial interior updates. A training sample is a pair $(\Omega, g)$ obtained as follows.

\paragraph{Geometry sampling.} Each training domain $\Omega = \bigcup_{k \in \mathcal{A}} C_k \subseteq \Omega_0$ is defined by an active set $\mathcal{A} \subseteq \{1, \ldots, 27\}$ of solid cells. We draw $\mathcal{A}$ uniformly from the set of admissible configurations
\begin{equation}
\mathcal{T} = \bigl\{\, \mathcal{A} \subseteq \{1, \ldots, 27\} : \mathcal{A} \neq \emptyset,\; \mathcal{A} \neq \{k_c\},\; \mathcal{A} \text{ is face-connected} \,\bigr\},
\end{equation}
where $k_c$ is the central cell. Here, \emph{face-connected} means that the graph $(\mathcal{A}, {\sim})$, with $k \sim k'$ if and only if $C_k$ and $C_{k'}$ share a full two-dimensional face, is connected. This guarantees that the solid is a single continuous component, while admitting arbitrary void configurations.

\begin{figure}[h]
    \centering   \includegraphics[width=1\linewidth]{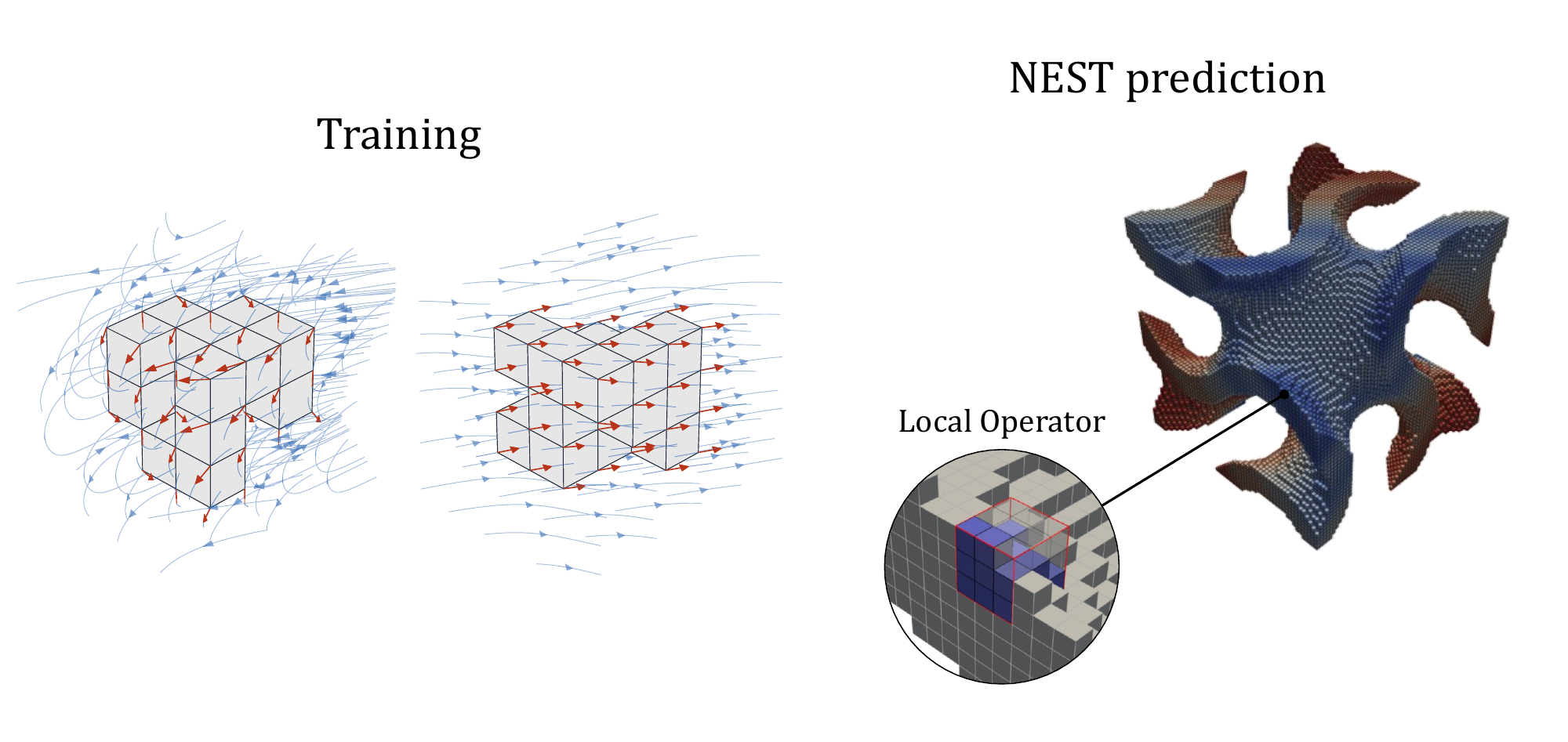}
    \caption{\textbf{Left:} Assignment of displacement boundary conditions. A continuous displacement field (represented by blue streamlines) is evaluated at the boundary nodes of the voxelized domain to prescribe Dirichlet boundary conditions (red vectors). \textbf{Right:} A close-up highlights the local operator tiling geometries during inference.}
    \label{fig:bcs}
\end{figure}

\paragraph{Boundary data sampling.} Given a sampled domain $\Omega$, we prescribe the Dirichlet datum $g$ on $\partial \Omega$ as a smooth, multiscale random field. Generalizing the procedure introduced in \cite{wu2026learning} to 3D, let $x = (x_1, x_2, x_3) \in \Omega$ denote material coordinates and define the projection $s(x) = (x_1 + x_2 + x_3)/3$ together with its rescaled counterpart $\tilde{s}(x) = (s(x) + \delta)/\zeta + s_0$. The components of $g$, for $i \in \{1,2,3\}$, are then
\begin{equation}
g_i(x) = \alpha \sum_{n=1}^{N} \frac{1}{(n+1)^{\kappa}} \Bigl[ a_n^{(i)} \cos\bigl( 2\pi n\, \tilde{s}(x) + b_n^{(i)} \bigr) + c_n^{(i)} \sin\bigl( 2\pi n\, \tilde{s}(x) + d_n^{(i)} \bigr) + e_n^{(i)} \Bigr].
\end{equation}
The amplitudes $a_n^{(i)}, c_n^{(i)} \sim \mathcal{U}[0.5, 1.0]$, phases $b_n^{(i)}, d_n^{(i)} \sim \mathcal{U}[\pi/4, 5\pi/4]$, and biases $e_n^{(i)} \sim \mathcal{U}[-0.25, 0.25]$ are sampled independently per component $i$ and per mode $n$; $N$ is the number of active modes, $\kappa$ a fixed spectral-decay exponent, and $\alpha$ a global magnitude. The phase shift $s_0 \sim \mathcal{U}[0,1]$, offset $\delta \sim \mathcal{U}[0, \zeta]$, and \emph{zoom} $\log \zeta \sim \mathcal{U}[\log \zeta_{\min}, \log \zeta_{\max}]$ are sampled once per domain. The pair $(\zeta, \delta)$ selects a window of length $1/\zeta$ from the field, so the patch sees boundary data consistent with what a $3\times3\times3$ tile would receive inside a $\zeta$-times-larger structure, therefore exposing the surrogate at training time to the multiscale gradients it must handle at Schwarz inference on generic domains. The field $g$ is evaluated at the boundary nodes of the FEM discretization of $\Omega$ and assigned as Dirichlet data there, as illustrated in Figure \ref{fig:bcs}.

\paragraph{Architecture.} We parameterize $\mathcal{S}_\theta$ as a Graph Neural Operator (GNO)~\cite{li2020neural,kovachki2023neural} on the patch graph $\mathcal{G}=(\mathcal{V}, \mathcal{E})$. The node set $\mathcal{V}$ consists of the corners of the $4 \times 4 \times 4$ reference grid that are incident to at least one cell of $\Omega$, and is partitioned into boundary nodes $\partial \mathcal{V}$ on $\partial \Omega$ and interior (active) nodes $\mathcal{V} \setminus \partial \mathcal{V}$. Edges $\mathcal{E}=\{(i,j) : 0 < \|x_i - x_j\| \le r\}$ connect node pairs within a fixed radius $r = 1.5\,h$, where $h$ is the cell edge length. The Dirichlet datum $g$, prescribed on $\partial \mathcal{V}$, is extended to every node by inverse-distance-squared interpolation \cite{wu2026learning},
\begin{equation}
  w_{ij} = \frac{\|x_i - x_j\|^{-2}}{\sum_{k \in \partial \mathcal{V}} \|x_i - x_k\|^{-2}},
\end{equation}

\begin{equation}
   \tilde{g}(x_i) = \sum_{j \in \partial \mathcal{V}} w_{ij}\, g(x_j) \;\; \forall i \in \mathcal{V} \setminus \partial \mathcal{V}, \quad
\tilde{g}(x_i) = g(x_i) \;\; \forall i \in \partial \mathcal{V},
\end{equation}

yielding the input feature $h_i^{(0)} = \tilde{g}(x_i) \in \mathbb{R}^3$ at every node. A lifting MLP maps $h_i^{(0)}$ to a hidden width $C$, after which $L$ kernel-integral layers update features by
\begin{equation}
h_i^{(\ell+1)} = h_i^{(\ell)} + \sigma\!\left( \mathrm{LN}\!\left( W^{(\ell)} h_i^{(\ell)} + \frac{1}{|\mathcal{N}(i)|} \sum_{j \in \mathcal{N}(i)} \kappa^{(\ell)}\!\bigl(x_j - x_i,\, h_i^{(\ell)}, h_j^{(\ell)}\bigr) \odot h_j^{(\ell)} \right) \right),
\end{equation}
where $\mathcal{N}(i)$ are the graph neighbours of $i$, $\kappa^{(\ell)}$ is an MLP producing per-channel multiplicative weights from the relative position and the features of the two incident nodes, $\sigma$ is the SiLU activation, and $\mathrm{LN}$ denotes layer normalization. A node-wise projection MLP then maps $h_i^{(L)}$ to the prediction $\hat{y}(x_i)$. We train two variants of $\mathcal{S}_\theta$ that share this architecture but differ in their output target: a \emph{displacement} surrogate with $\hat{y}(x_i) = \hat{u}(x_i) \in \mathbb{R}^3$, and a \emph{gradient} surrogate with $\hat{y}(x_i) = \widehat{\nabla u}_{11}(x_i) \in \mathbb{R}$, the same logic can be potentially extended to the other gradient components. The displacement gradient can be then used to compute strain and stress fields. Predicting $\nabla u$ with a dedicated surrogate, rather than recovering it by differentiating $\hat{u}$, is empirically less sensitive to small displacement errors.

\paragraph{Training.} \label{par:training}We generate $15{,}000$ samples and split them $90$/$10$ into train and validation. Both surrogates are trained for $100$ epochs with Adam \cite{kingma2014adam} at learning rate $10^{-4}$ and the checkpoint with the lowest validation loss is retained. Inputs and outputs are channel-wise $z$-normalized using statistics computed on the training split. The gradient surrogate is trained with the MSE on its scalar output. The displacement surrogate is trained with the composite loss $\mathcal{L} = \mathrm{MSE}(\hat{u}, u) + \mathrm{MSE}(\nabla \hat{u}, \nabla u)$. This term acts as a gradient-consistency regularizer, indeed, at Schwarz inference, the converged displacement prediction of each patch is fed into the gradient surrogate to produce the $\nabla u$.

\subsection{Schwarz Method}
\label{subsec:schwarz}
Having trained the local surrogate $\mathcal{S}_\theta$, we approximate the global solution on an arbitrary domain via an overlapping additive Schwarz method, employing $\mathcal{S}_\theta$ as the local solver. We cover $\Omega$ with overlapping $3 \times 3 \times 3$ patches $\{\Omega_p\}_{p=1}^{P}$, each matching the canonical training geometry as illustrated in Figure \ref{fig:bcs}. We write the boundary of each patch as $\partial \Omega_p = \Gamma_p^{\mathrm{ext}} \cup \Gamma_p^{\mathrm{int}}$. Here, $\Gamma_p^{\mathrm{ext}} = \partial \Omega_p \cap \partial \Omega$ inherits the macroscopic Dirichlet datum, and $\Gamma_p^{\mathrm{int}}$ is the interface with neighbouring patches. Let $\mathcal{V}$ denote the active nodes of $\Omega$, and let $\mathcal{V}_p \subseteq \mathcal{V}$ denote those of patch $p$. Finally, let $\{\chi_p\}_{p=1}^{P}$ be a partition of unity over $\mathcal{V}$ such that each $\chi_p$ vanishes identically outside $\mathcal{V}_p$.

Starting from an initial iterate $u^{(0)}$ that matches $g$ on $\partial \Omega$, each Schwarz sweep $n \to n+1$ proceeds, in parallel for every patch, by assembling local Dirichlet data and invoking the displacement surrogate,
\begin{equation}
g_p^{(n)} = \begin{cases} g & \text{on } \Gamma_p^{\mathrm{ext}}, \\ u^{(n)}\big|_{\Gamma_p^{\mathrm{int}}} & \text{on } \Gamma_p^{\mathrm{int}}, \end{cases}
\qquad
\hat u_p^{(n+1)} = \mathcal{S}_\theta(\Omega_p, g_p^{(n)}),
\label{eq:schwarz_local}
\end{equation}
followed by partition-of-unity assembly into a global field,
\begin{equation}
u^{(n+1)} = \sum_{p=1}^{P} \chi_p\, \hat u_p^{(n+1)}, \qquad u^{(n+1)}\big|_{\partial \Omega} = g.
\label{eq:schwarz_assembly}
\end{equation}
Iteration stops once the relative change on interior active nodes,
\begin{equation}
\Delta^{(n+1)} = \frac{\|u^{(n+1)} - u^{(n)}\|_{2,\,\mathcal{V}\setminus \partial\Omega}}{\|u^{(n+1)}\|_{2,\,\mathcal{V}\setminus \partial\Omega}},
\label{eq:delta}
\end{equation}
falls below a prescribed tolerance $\varepsilon$. Algorithm details can be found in Appendix \ref{app:pseudocode}.

Once the displacement iteration has converged to $u^\star$, we recover $\nabla u$ on $\Omega$ by a single non-iterative pass of the gradient surrogate. For each patch $p$, we extract the converged displacement $u^\star\big|_{\partial \Omega_p}$ on the patch boundary and feed it as Dirichlet datum to a dedicated GNO trained, as described above, to predict one component of $\nabla u$. Repeating the pass per component and assembling the patchwise predictions through the partition of unity $\{\chi_p\}$ yields the global gradient field.

\section{Experiments}
\label{sec:experiments}

We design the experiments to evaluate three claims: 
(i) NEST trained on $3 \times 3 \times 3$ patches generalizes to much larger, unseen, geometrically distinct domains without re-training, similarly to FEM; 
(ii) once trained, contrary to FEM, inference cost is largely independent of the complexity of the constitutive model; and 
(iii) local patch training provides a paradigm shift: from global training on full-domain solutions to reusable locally trained neural surrogates.

\begin{figure}[htbp]
    \centering
    \includegraphics[width=1\textwidth]{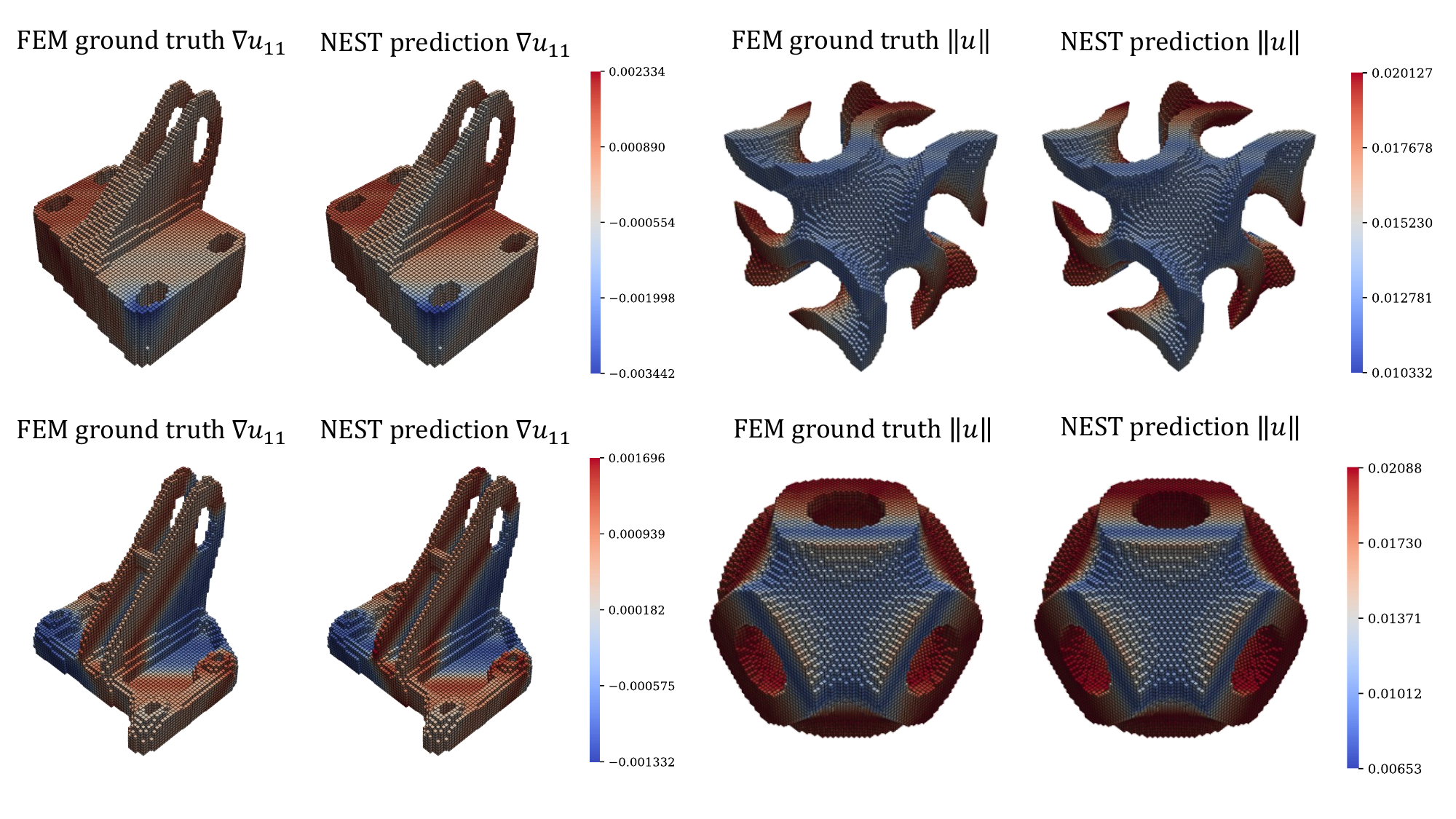}
    \caption{NEST vs. FEM on unseen macro-scale test geometries. The test set contains two SimJEB jet-engine brackets and two TPMS structures, voxelized at multiple resolutions, here shown at $n = 60$. NEST is trained only on random $3 \times 3 \times 3$ local patches and is not fine-tuned on these geometries.}
    \label{fig:test_dataset}
\end{figure}

All experiments use the displacement and gradient surrogates described in Paragraph~\ref{par:training}. The local models are trained once and then frozen for all macro-scale evaluations. The test set contains four macro-scale geometries: two triply periodic minimal surface (TPMS) structures and two jet-engine brackets from the SimJEB dataset~\cite{whalen2021simjeb}. These geometries are not used during local training. Each geometry is voxelized at multiple resolutions $n \times n \times n$, where $n$ denotes the number of voxels per axis. The results for $n = 60$ in Figure \ref{fig:test_dataset} highlight the generalization capabilities of NEST. Unless otherwise stated, NEST is run with the overlapping Schwarz iteration described in Section~4.2, using a convergence tolerance of $\varepsilon = 10^{-5}$ for the relative iterate change in Eq.~\eqref{eq:delta}. An ablation isolating the role of Schwarz coupling is reported in Appendix~\ref{subsec:schwarz_ablation}, while running time is detailed in Appendix \ref{app:runtime}. Reference solutions are computed with the FEniCS finite-element solver. 

\subsection{Accuracy on unseen macro-scale geometries}
\label{subsec:accuracy}

Table~\ref{tab:error_results} reports max-normalized mean-squared errors of NEST on unseen SimJEB bracket and TPMS geometries at increasing voxel resolutions, for displacement and for the $11$-component of the displacement gradient; full error definitions are given in Appendix~\ref{app:metrics}. A single local model trained on $3 \times 3 \times 3$ patches is used for all geometries and resolutions. No macro-scale data are used for training or fine-tuning. When multiple geometries belong to the same class, the reported values are averaged within that class.

\begin{table}[htbp]
\centering
\caption{Accuracy of NEST on unseen macro-scale geometries. A single local model trained on $3 \times 3 \times 3$ patches is reused for all test geometries and resolutions.}
\label{tab:error_results}
\begin{tabular}{c c c c c}
\toprule
 & \multicolumn{2}{c}{\textbf{SimJEB brackets}} & \multicolumn{2}{c}{\textbf{TPMS}} \\
\cmidrule(lr){2-3} \cmidrule(lr){4-5}
\textbf{Resolution} & $e_{u}^{\mathrm{maxMSE}}$ & $e_{\nabla u_{11}}^{\mathrm{maxMSE}}$ & $e_{u}^{\mathrm{maxMSE}}$ & $e_{\nabla u_{11}}^{\mathrm{maxMSE}}$ \\
\midrule
30 & $8.147 \times 10^{-3}$ & $3.102 \times 10^{-2}$ & $5.574 \times 10^{-3}$ & $3.376 \times 10^{-3}$ \\
45 & $6.670 \times 10^{-3}$ & $2.566 \times 10^{-2}$ & $2.539 \times 10^{-3}$ & $5.155 \times 10^{-3}$ \\
60 & $6.929 \times 10^{-3}$ & $2.096 \times 10^{-2}$ & $1.715 \times 10^{-3}$ & $5.997 \times 10^{-3}$ \\
\bottomrule
\end{tabular}
\end{table}

NEST maintains stable displacement accuracy as the macro-scale resolution increases, despite being trained only on local voxel patches. The displacement error remains below $10^{-2}$ across all tested geometries and resolutions. Errors are higher on SimJEB brackets than on TPMS structures, which is expected because the bracket geometries contain sharper geometric features and less regular local topology. Nevertheless, the same local model remains applicable across both geometry classes, supporting the central premise that local neural building blocks can be composed through Schwarz iteration to solve geometrically distinct domains.

\subsection{Comparison with a favorable global neural-operator baseline}
\label{subsec:baseline}

We next compare NEST with a global neural-operator baseline in an intentionally favorable setting for the baseline. 
A standard global neural operator is trained to map full macro-scale geometries and boundary data directly to full solution fields. 
In practice, such models are known to be highly sensitive to the geometry distribution used for training: when trained on only a small number of macro-scale geometries and evaluated on substantially different unseen geometries, performance typically deteriorates dramatically. 
For this reason, we do not use the global baseline as a strict out-of-distribution geometry-generalization test. 
Instead, we ask a more conservative question: can NEST, trained only on local $3 \times 3 \times 3$ patches and never on macro-scale solutions, match a global neural operator trained directly on the target macro-scale geometry class?

We use Transolver~\cite{wu2024transolver} as a representative global neural-operator baseline. 
The comparison is deliberately favorable to Transolver: the model is trained on macro-scale TPMS and SimJEB geometries at resolution $60^3$, using full-domain FEM solutions. 
In contrast, NEST is trained only once on local voxel patches and is reused without fine-tuning on all macro-scale geometries and resolutions. 
Thus, while Transolver learns a geometry-specific full-domain surrogate, NEST learns a reusable local solver that is composed through Schwarz iteration at inference time. Transolver was trained for $1000$ epochs with an initial learning rate of $10^{-3}$, halved every $200$ epochs. 
On the TPMS geometries at resolution $60^3$, Transolver reaches a displacement error of $1.740 \times 10^{-3}$ and a gradient-component error of $9.392 \times 10^{-3}$ for $\nabla u_{11}$. 
On the same setting, NEST reaches a displacement error of $1.715 \times 10^{-3}$ and a gradient-component error of $5.997 \times 10^{-3}$. 
Therefore, NEST matches the displacement accuracy of a global neural operator trained directly on macro-scale data, while producing a more accurate estimate of the displacement-gradient component.

This comparison should be interpreted as an upper-bound-style baseline rather than as a standard out-of-distribution benchmark for global neural operators. 
Its purpose is to test whether local neural-Schwarz composition can be competitive with a global surrogate that has access to far more task-specific information during training. 
The result is notable because NEST does not require macro-scale training data, does not observe the target geometries during training, and reuses the same local model across all tested geometry classes and resolutions.

\subsection{Reusable local training versus geometry-specific global training}
\label{subsec:data_efficiency}

The previous comparison evaluates predictive accuracy, but it does not by itself capture the main data advantage of NEST. 
The key distinction is not only the cost of generating one training solve, but the \emph{type} of information contained in the training set. 
Global neural operators are trained on full-domain examples tied to a particular macro-scale geometry distribution. 
Consequently, when the target geometry class, resolution, or boundary-condition structure changes substantially, new macro-scale FEM data may be required. 
In contrast, NEST is trained on local $3 \times 3 \times 3$ patch problems. 
These local problems are not associated with a specific macro-scale geometry; they define reusable physical building blocks that can be composed through Schwarz iteration on different domains.

To make this comparison concrete, we compare NEST and global neural-operator baselines under a similar data-generation budget. 
NEST is trained on $15{,}000$ local patch solves, corresponding to $2.412\times 10^6$ total degrees of freedom and $8.880\times 10^3$ seconds of data-generation time. 
The global neural-operator baselines are trained on full macro-scale FEM solutions from TPMS and SimJEB geometries, corresponding to $2.575\times 10^6$ total degrees of freedom and $9.062\times 10^3$ seconds of data-generation time. 
Thus, the total amount of generated FEM data and wall-clock data-generation time are comparable across the two settings, but the resulting training sets have fundamentally different reuse properties.

\begin{table}[htbp]
\centering
\caption{Training-data comparison between local NEST training and global neural-operator training. The total generated degrees of freedom and data-generation wall time are comparable, but NEST uses local patch problems that can be reused across macro-scale geometries and resolutions, whereas the global baselines use full-domain examples tied to the target geometry families.}
\label{tab:data_efficiency}
\begin{tabular}{l c c c c}
\toprule
\textbf{Method} & \textbf{Training domains} & \textbf{\# FEM solves} & \textbf{Tot DoFs} & \textbf{Data-gen. time [s]} \\
\midrule
NEST & Local $3 \times 3 \times 3$ patches & $15{,}000$ & $2.412\times 10^6$ & $8.880 \times 10^3$\\
NOs & Macro-scale geometries & 13 & $2.575\times 10^6$ & $9.062\times 10^3$ \\
\bottomrule
\end{tabular}
\end{table}

For the global neural-operator baselines, we train Transolver and GNO on a macro-scale dataset consisting of $7$ TPMS geometries, obtained from linear combinations of $8$ primitive TPMS equations \cite{al2021mslattice,perroni2025data}, and $6$ SimJEB bracket geometries. 
The dataset is split $90/10$ into training and validation sets. 
For each architecture, two separate models are trained, one for the displacement field $u$ and one for the gradient component $\nabla u_{11}$. 
The models are trained for $500$ epochs with an initial learning rate of $10^{-3}$, halved every $100$ epochs. 
Errors are evaluated on the test geometries shown in Figure~\ref{fig:test_dataset} at resolution $60^3$.

\begin{table}[htbp]
\centering
\caption{Accuracy comparison between global neural-operator baselines and NEST at resolution $60^3$. The global baselines are trained on macro-scale data from the target geometry families, whereas NEST is trained only on local $3 \times 3 \times 3$ patch problems and reused without macro-scale fine-tuning.}
\label{tab:error_comparison_60}
\begin{tabular}{c c c c c c} 
\toprule
 & & \multicolumn{2}{c}{\textbf{Brackets}} & \multicolumn{2}{c}{\textbf{TPMS}} \\
\cmidrule(lr){3-4} \cmidrule(lr){5-6}
\textbf{Model} & \textbf{Parameters $u$ model} & $e_{\|u\|}^{\mathrm{maxMSE}}$ & $ e_{\nabla u_{11}}^{\mathrm{maxMSE}}$ & $e_{\|u\|}^{\mathrm{maxMSE}}$ & $ e_{\nabla u_{11}}^{\mathrm{maxMSE}}$ \\
\midrule
Transolver & $3.885\times 10^5$ & 1.457E-02 & 3.932E-02 & 5.156E-02 & 1.286E-01 \\
GNO & $2.011\times 10^5$ & 8.601E-03 & 3.433E-02 & 2.449E-01 & 4.405E-02 \\
NEST & $7.661\times 10^4 $ & \textbf{6.929E-03} & \textbf{2.096E-02} & \textbf{1.715E-03} & \textbf{5.997E-03} \\
\bottomrule
\end{tabular}
\label{table:err}
\end{table}

Under a comparable data-generation budget, NEST achieves the lowest errors across both geometry classes and both reported quantities. 
More importantly, the comparison highlights a qualitative difference in how the generated data are used. 
The global neural operators consume macro-scale solutions associated with specific geometry families, so their training data must cover the target distribution of full domains. 
NEST instead consumes local physical solves that are independent of any particular macro-scale geometry. 
The same trained local model is then reused across TPMS and SimJEB structures and across multiple resolutions through Schwarz assembly. 
Thus, the advantage of NEST is best understood as \emph{reusable local training}: comparable data-generation effort produces a composable local solver rather than a geometry-specific global surrogate.

\section{Conclusion}

We introduced NEST, a neural-Schwarz framework that learns local solvers on minimal voxel patches and composes them through overlapping Schwarz iteration to solve PDEs on larger, unseen domains. By separating local physics learning from global assembly, NEST avoids geometry-specific full-domain training and enables reuse across sizes, shapes, and boundary conditions. On nonlinear 3D elasticity, the same locally trained model generalizes across geometries and surpasses state-of-the-art neural-operator baselines trained on macro-scale data, even under favorable evaluation for those baselines. Once trained, NEST also keeps inference largely independent of PDE nonlinearities. These results support local-to-global composition as a path toward reusable learned PDE solvers. Future work will improve Schwarz convergence and extend NEST to richer physics and boundary conditions.

\section{Data and Code Availability}
The datasets and source code supporting the findings of this work will be made publicly available upon acceptance of the manuscript.

\section{Acknowledgements}
This work was funded by the UKRI Engineering and Physical Sciences Research Council (EPSRC), and Tata Steel Research and Innovation Ltd (TSRIL), UK, through an Industrial Cooperative Awards in Science and Engineering (iCase) award, number 220109, to Paolo Secchi. 
Furthermore, the authors gratefully acknowledge Marco Benedetti for his valuable insights and constructive discussions.

\bibliographystyle{unsrtnat} 
\bibliography{biblio} 

@article{ma2026learning,
  title={Learning data-efficient and generalizable neural operators via fundamental physics knowledge},
  author={Ma, Siying and Zadeh, Mehrdad M and Soroco, Mauricio and Chen, Wuyang and Cao, Jiguo and Ganesh, Vijay},
  journal={arXiv preprint arXiv:2602.15184},
  year={2026}
}

@article{li2020neural,
  title={Neural operator: Graph kernel network for partial differential equations},
  author={Li, Zongyi and Kovachki, Nikola and Azizzadenesheli, Kamyar and Liu, Burigede and Bhattacharya, Kaushik and Stuart, Andrew and Anandkumar, Anima},
  journal={arXiv preprint arXiv:2003.03485},
  year={2020}
}

@book{logg2012automated,
  title={Automated solution of differential equations by the finite element method: The FEniCS book},
  author={Logg, Anders and Mardal, Kent-Andre and Wells, Garth},
  volume={84},
  year={2012},
  publisher={Springer Science \& Business Media}
}

@inproceedings{whalen2021simjeb,
  title={SimJEB: simulated jet engine bracket dataset},
  author={Whalen, Eamon and Beyene, Azariah and Mueller, Caitlin},
  booktitle={Computer Graphics Forum},
  volume={40},
  pages={9--17},
  year={2021},
  organization={Wiley Online Library}
}

@article{li2020fourier,
  title={Fourier neural operator for parametric partial differential equations},
  author={Li, Zongyi and Kovachki, Nikola and Azizzadenesheli, Kamyar and Liu, Burigede and Bhattacharya, Kaushik and Stuart, Andrew and Anandkumar, Anima},
  journal={arXiv preprint arXiv:2010.08895},
  year={2020}
}

@article{raissi2019physics,
  title={Physics-informed neural networks: A deep learning framework for solving forward and inverse problems involving nonlinear partial differential equations},
  author={Raissi, Maziar and Perdikaris, Paris and Karniadakis, George E},
  journal={Journal of Computational physics},
  volume={378},
  pages={686--707},
  year={2019},
  publisher={Elsevier}
}

@article{li2024physics,
  title={Physics-Informed Neural Operator for Learning Partial Differential Equations},
  author={Li, Zongyi and Zheng, Hongkai and Kovachki, Nikola and Jin, David and Chen, Haoxuan and Liu, Burigede and Azizzadenesheli, Kamyar and Anandkumar, Anima},
  journal={ACM/IMS Journal of Data Science},
  volume={1},
  number={3},
  articleno={9},
  pages={1--27},
  year={2024},
  doi={10.1145/3648506},
  url={https://doi.org/10.1145/3648506}
}

@article{lu2021learning,
  title={Learning nonlinear operators via {DeepONet} based on the universal approximation theorem of operators},
  author={Lu, Lu and Jin, Pengzhan and Pang, Guofei and Zhang, Zhongqiang and Karniadakis, George Em},
  journal={Nature Machine Intelligence},
  volume={3},
  number={3},
  pages={218--229},
  year={2021},
  doi={10.1038/s42256-021-00302-5},
  url={https://doi.org/10.1038/s42256-021-00302-5}
}

@article{kovachki2023neural,
  title={Neural Operator: Learning Maps Between Function Spaces With Applications to {PDEs}},
  author={Kovachki, Nikola and Li, Zongyi and Liu, Burigede and Azizzadenesheli, Kamyar and Bhattacharya, Kaushik and Stuart, Andrew and Anandkumar, Anima},
  journal={Journal of Machine Learning Research},
  volume={24},
  number={89},
  pages={1--97},
  year={2023},
  url={https://jmlr.org/papers/v24/21-1524.html}
}

@inproceedings{pfaff2021learning,
  title={Learning Mesh-Based Simulation with Graph Networks},
  author={Pfaff, Tobias and Fortunato, Meire and Sanchez-Gonzalez, Alvaro and Battaglia, Peter W.},
  booktitle={International Conference on Learning Representations},
  year={2021},
  url={https://openreview.net/forum?id=roNqYL0_XP}
}

@inproceedings{li2023geometry,
  title={Geometry-Informed Neural Operator for Large-Scale 3D {PDEs}},
  author={Li, Zongyi and Kovachki, Nikola and Choy, Christopher and Li, Boyi and Kossaifi, Jean and Otta, Shourya and Nabian, Mohammad Amin and Stadler, Maximilian and Hundt, Christian and Azizzadenesheli, Kamyar and Anandkumar, Animashree},
  booktitle={Advances in Neural Information Processing Systems},
  volume={36},
  year={2023},
  url={https://proceedings.neurips.cc/paper_files/paper/2023/hash/70518ea42831f02afc3a2828993935ad-Abstract-Conference.html}
}

@inproceedings{wu2024transolver,
  title={Transolver: A Fast Transformer Solver for {PDE}s on General Geometries},
  author={Wu, Haixu and Luo, Huakun and Wang, Haowen and Wang, Jianmin and Long, Mingsheng},
  booktitle={Proceedings of the 41st International Conference on Machine Learning},
  pages={53681--53705},
  year={2024},
  volume={235},
  series={Proceedings of Machine Learning Research},
  publisher={PMLR},
  url={https://proceedings.mlr.press/v235/wu24r.html}
}

@inproceedings{bryutkin2024hamlet,
  title={{HAMLET}: Graph Transformer Neural Operator for Partial Differential Equations},
  author={Bryutkin, Andrey and Huang, Jiahao and Deng, Zhongying and Yang, Guang and Sch{\"o}nlieb, Carola-Bibiane and Aviles-Rivero, Angelica I.},
  booktitle={Proceedings of the 41st International Conference on Machine Learning},
  pages={4624--4641},
  year={2024},
  volume={235},
  series={Proceedings of Machine Learning Research},
  publisher={PMLR},
  url={https://proceedings.mlr.press/v235/bryutkin24a.html}
}

@book{dolean2015introduction,
  title={An Introduction to Domain Decomposition Methods: Algorithms, Theory, and Parallel Implementation},
  author={Dolean, Victorita and Jolivet, Pierre and Nataf, Fr{\'e}d{\'e}ric},
  series={Other Titles in Applied Mathematics},
  volume={144},
  publisher={Society for Industrial and Applied Mathematics},
  address={Philadelphia, PA},
  year={2015},
  doi={10.1137/1.9781611974065},
  url={https://doi.org/10.1137/1.9781611974065}
}

@inproceedings{taghibakhshi2022learning,
  title={Learning Interface Conditions in Domain Decomposition Solvers},
  author={Taghibakhshi, Ali and Nytko, Nicolas and Zaman, Tareq Uz and MacLachlan, Scott and Olson, Luke and West, Matthew},
  booktitle={Advances in Neural Information Processing Systems},
  volume={35},
  year={2022},
  url={https://proceedings.neurips.cc/paper_files/paper/2022/hash/2f8928efe957139e9c0efc98f173f4be-Abstract-Conference.html}
}

@article{wu2026learning,
  title={A learning-based domain decomposition method},
  author={Wu, Rui and Kovachki, Nikola and Liu, Burigede},
  journal={Computer Methods in Applied Mechanics and Engineering},
  volume={453},
  pages={118799},
  year={2026},
  publisher={Elsevier}
}

@article{huang2025operator,
  title={Operator Learning with Domain Decomposition for Geometry Generalization in PDE Solving},
  author={Huang, Jianing and Zhang, Kaixuan and Wu, Youjia and Cheng, Ze},
  journal={arXiv preprint arXiv:2504.00510},
  year={2025}
}

@article{luo2025transolver++,
  title={Transolver++: An accurate neural solver for pdes on million-scale geometries},
  author={Luo, Huakun and Wu, Haixu and Zhou, Hang and Xing, Lanxiang and Di, Yichen and Wang, Jianmin and Long, Mingsheng},
  journal={arXiv preprint arXiv:2502.02414},
  year={2025}
}

@article{brown2020language,
  title={Language models are few-shot learners},
  author={Brown, Tom and Mann, Benjamin and Ryder, Nick and Subbiah, Melanie and Kaplan, Jared D and Dhariwal, Prafulla and Neelakantan, Arvind and Shyam, Pranav and Sastry, Girish and Askell, Amanda and others},
  journal={Advances in neural information processing systems},
  volume={33},
  pages={1877--1901},
  year={2020}
}

@book{toselli2004domain,
  title={Domain decomposition methods-algorithms and theory},
  author={Toselli, Andrea and Widlund, Olof},
  volume={34},
  year={2004},
  publisher={Springer Science \& Business Media}
}

@article{ouyang2026noem,
  title={NOEM: efficient and scalable finite element method enabled by reusable neural operators},
  author={Ouyang, Weihang and Shin, Yeonjong and Liu, Si-Wei and Lu, Lu},
  journal={Nature Computational Science},
  volume={6},
  number={4},
  pages={417--429},
  year={2026},
  publisher={Nature Publishing Group US New York}
}

@article{subramanian2023towards,
  title={Towards foundation models for scientific machine learning: Characterizing scaling and transfer behavior},
  author={Subramanian, Shashank and Harrington, Peter and Keutzer, Kurt and Bhimji, Wahid and Morozov, Dmitriy and Mahoney, Michael W and Gholami, Amir},
  journal={Advances in Neural Information Processing Systems},
  volume={36},
  pages={71242--71262},
  year={2023}
}

@article{mccabe2024multiple,
  title={Multiple physics pretraining for spatiotemporal surrogate models},
  author={McCabe, Michael and R{\'e}galdo-Saint Blancard, Bruno and Parker, Liam and Ohana, Ruben and Cranmer, Miles and Bietti, Alberto and Eickenberg, Michael and Golkar, Siavash and Krawezik, Geraud and Lanusse, Francois and others},
  journal={Advances in Neural Information Processing Systems},
  volume={37},
  pages={119301--119335},
  year={2024}
}

@article{karniadakis2021physics,
  title={Physics-informed machine learning},
  author={Karniadakis, George Em and Kevrekidis, Ioannis G and Lu, Lu and Perdikaris, Paris and Wang, Sifan and Yang, Liu},
  journal={Nature Reviews Physics},
  volume={3},
  number={6},
  pages={422--440},
  year={2021},
  publisher={Nature Publishing Group UK London}
}

@article{li2023fourier,
  title={Fourier neural operator with learned deformations for pdes on general geometries},
  author={Li, Zongyi and Huang, Daniel Zhengyu and Liu, Burigede and Anandkumar, Anima},
  journal={Journal of Machine Learning Research},
  volume={24},
  number={388},
  pages={1--26},
  year={2023}
}

@article{li2025geometric,
  title={Geometric operator learning with optimal transport},
  author={Li, Xinyi and Li, Zongyi and Kovachki, Nikola and Anandkumar, Anima},
  journal={arXiv preprint arXiv:2507.20065},
  year={2025}
}

@article{al2021mslattice,
  title={MSLattice: A free software for generating uniform and graded lattices based on triply periodic minimal surfaces},
  author={Al-Ketan, Oraib and Abu Al-Rub, Rashid K},
  journal={Material Design \& Processing Communications},
  volume={3},
  number={6},
  pages={e205},
  year={2021},
  publisher={Wiley Online Library}
}

@inproceedings{perroni2025data,
  title={Data-Efficient Discovery of Hyperelastic TPMS Metamaterials with Extreme Energy Dissipation},
  author={Perroni-Scharf, Maxine and Ferguson, Zachary and Butruille, Thomas and Portela, Carlos and Konakovi{\'c} Lukovi{\'c}, Mina},
  booktitle={Proceedings of the Special Interest Group on Computer Graphics and Interactive Techniques Conference Conference Papers},
  pages={1--12},
  year={2025}
}

@article{vaswani2017attention,
  title={Attention is all you need},
  author={Vaswani, Ashish and Shazeer, Noam and Parmar, Niki and Uszkoreit, Jakob and Jones, Llion and Gomez, Aidan N and Kaiser, {\L}ukasz and Polosukhin, Illia},
  journal={Advances in neural information processing systems},
  volume={30},
  year={2017}
}

@article{kingma2014adam,
  title={Adam: A method for stochastic optimization},
  author={Kingma, Diederik P and Ba, Jimmy},
  journal={arXiv preprint arXiv:1412.6980},
  year={2014}
}

@article{tierz2025feasibility,
  title={On the feasibility of foundational models for the simulation of physical phenomena},
  author={Tierz, Alicia and Iparraguirre, Mikel M and Alfaro, Ic{\'\i}ar and Gonz{\'a}lez, David and Chinesta, Francisco and Cueto, El{\'\i}as},
  journal={International Journal for Numerical Methods in Engineering},
  volume={126},
  number={6},
  pages={e70027},
  year={2025},
  publisher={Wiley Online Library}
}

@article{masliaev2025towards,
  title={Towards Universal Neural Operators through Multiphysics Pretraining},
  author={Masliaev, Mikhail and Gusarov, Dmitry and Markov, Ilya and Hvatov, Alexander},
  journal={arXiv preprint arXiv:2511.10829},
  year={2025}
}

@article{zhu2025generalizing,
  title={Generalizing PDE Emulation with Equation-Aware Neural Operators},
  author={Zhu, Qian-Ze and Raccuglia, Paul and Brenner, Michael P},
  journal={arXiv preprint arXiv:2511.09729},
  year={2025}
}

@book{mathew2008domain,
  title={Domain decomposition methods for the numerical solution of partial differential equations},
  author={Mathew, Tarek Poonithara Abraham},
  year={2008},
  publisher={Springer}
}

@article{maurizi2025designing,
  title={Designing metamaterials with programmable nonlinear responses and geometric constraints in graph space},
  author={Maurizi, Marco and Xu, Derek and Wang, Yu-Tong and Yao, Desheng and Hahn, David and Oudich, Mourad and Satpati, Anish and Bauchy, Mathieu and Wang, Wei and Sun, Yizhou and others},
  journal={Nature Machine Intelligence},
  volume={7},
  number={7},
  pages={1023--1036},
  year={2025},
  publisher={Nature Publishing Group UK London}
}

\newpage
\appendix
\section*{Appendix}
\section{Error Metric}
\label{app:metrics}
The following error metrics have been employed:
\begin{equation}
e_{\|u\|}^{\mathrm{maxMSE}} = \frac{1}{N} \sum_{i=1}^N \frac{\left\|\hat{u}\left(x_i\right)-u\left(x_i\right)\right\|^2}{\max_j\left(\left\|u\left(x_j\right)\right\|^2\right)} \quad e_{\nabla u_{11}}^{\mathrm{maxMSE}} = \frac{1}{N} \sum_{i=1}^N \frac{\left(\widehat{\nabla u}_{11}\left(x_i\right)-\nabla u_{11}\left(x_i\right)\right)^2}{\max_j\left(\nabla u_{11}\left(x_j\right)^2\right)}
\label{eq:err}
\end{equation}
where $N$ is the total number of points in the test dataset. Errors are normalized by the maximum value to avoid division by zero, as $\nabla u$ is close to zero at certain points within the domain.

\section{Role of Schwarz coupling}
\label{subsec:schwarz_ablation}

To isolate the contribution of Schwarz iteration, we compare the full NEST solver against a one-shot local tiling baseline in Table \ref{tab:schwarz_ablation}. The one-shot baseline applies the trained local model independently on all patches and assembles the predictions with the same partition of unity, but does not iterate to enforce interface consistency.

\begin{table}[htbp]
\centering
\caption{Ablation of Schwarz coupling. The one-shot baseline uses the same trained local model and the same patch decomposition as NEST, but removes iterative interface coupling. The relative error on Gyroid TPMS with resolution 60 is evaluated.}
\label{tab:schwarz_ablation}
\begin{tabular}{l c c c}
\toprule
\textbf{Method} & \textbf{Resolution} & $e_{u}^{\mathrm{maxMSE}}$ & $e_{\nabla u_{11}}^{\mathrm{maxMSE}}$ \\
\midrule
One-shot local tiling & $60$ & $5.711\times 10^{-1}$ & $8.539\times 10^{-1}$ \\
NEST with Schwarz & $60$ & $1.424\times 10^{-3} $ & $5.593 \times 10^{-3}$ \\
\bottomrule
\end{tabular}
\end{table}

This ablation is important because it tests whether global accuracy comes from the local neural predictor alone or from the combination of local prediction and iterative Schwarz consistency.

\section{Resolution scaling and runtime}
\label{app:runtime}

We next evaluate the computational cost of NEST inference. Table~\ref{tab:speed} compares the wall-clock time of the proposed neural-Schwarz prediction pipeline against the FEniCS solver used to generate reference solutions. NEST inference is executed on a single RTX 6000 GPU, while FEM simulations are executed on a single CPU core. This comparison is therefore not intended as a hardware-normalized benchmark against optimized FEM solvers; rather, it quantifies the practical wall-clock speedup of the current NEST implementation relative to the FEM ground-truth generator used in our experiments.

\begin{table}[htbp]
    \centering
    \caption{Wall-clock runtime comparison on a Gyroid TPMS geometry. NEST is run on a single RTX 6000 GPU, while FEM is run on a single CPU core.}
    \label{tab:speed}
    \begin{tabular}{c c c c c c}
        \toprule
        \textbf{Resolution} & \textbf{Iterations} & \textbf{FEM (s)} & \textbf{NEST total (s)} & \textbf{NEST / iter. (s)} & \textbf{Speedup} \\
        \midrule
        15 & 4  & 29.90  & 0.27   & 0.07 & $110.7\times$ \\
        30 & 14 & 166.70 & 9.43   & 0.67 & $17.7\times$ \\
        45 & 29 & 459.60 & 57.81  & 1.99 & $8.0\times$ \\
        60 & 61 & 811.15 & 257.55 & 4.22 & $3.2\times$ \\
        \bottomrule
    \end{tabular}
\end{table}

NEST provides substantial wall-clock speedups at all tested resolutions. The speedup decreases with increasing resolution because the number of Schwarz iterations grows from $4$ iterations at resolution $15^3$ to $61$ iterations at resolution $60^3$. This behavior is consistent with one-level Schwarz methods, where global information propagates through local overlap communication. Thus, the current implementation should not be interpreted as demonstrating total $O(N)$ runtime. Rather, the per-iteration cost is expected to scale approximately linearly with the number of active patches/nodes, while the total runtime also depends on the number of Schwarz iterations required for global convergence.

To directly test this scaling claim, we decompose the total NEST runtime as
\begin{equation}
T_{\mathrm{NEST}} \approx N_{\mathrm{it}} \, T_{\mathrm{sweep}},
\qquad
T_{\mathrm{sweep}} = O(P^{\alpha}),
\label{eq:tnest_scaling}
\end{equation}
where $P$ is the number of active patches, $N_{\mathrm{it}}$ is the number of Schwarz iterations, and $\alpha$ is the per-sweep scaling exponent. The expected behavior $T_{\mathrm{sweep}} = O(P)$ corresponds to $\alpha = 1$. We estimate $\alpha$ empirically between consecutive resolutions as
\begin{equation}
\alpha_{i \to i+1} = \frac{\log\bigl(T_{i+1}/T_i\bigr)}{\log\bigl(P_{i+1}/P_i\bigr)},
\label{eq:alpha_empirical}
\end{equation}
where $T_i$ and $P_i$ denote the per-iteration runtime and number of active patches at resolution $i$. Table~\ref{tab:scaling} reports the number of active nodes, number of active patches, per-iteration runtime, and the resulting empirical exponent.

\begin{table}[htbp]
\centering
\caption{Empirical scaling of one NEST Schwarz sweep on a Gyroid TPMS geometry across resolutions. The empirical exponent $\alpha$ is computed pairwise between consecutive resolutions; values close to $1$ indicate the predicted $O(P)$ per-sweep cost.}
\label{tab:scaling}
\begin{tabular}{c c c c c}
\toprule
\textbf{Resolution} & \textbf{Active nodes} & \textbf{Active patches} & \textbf{NEST / iter. (s)} & \textbf{Empirical exponent $\alpha$} \\
\midrule
15 & $2.42\times 10^3$ & $1.11\times 10^3$ & 0.07 & --   \\
30 & $8.60\times 10^3$ & $1.37\times 10^4$ & 0.67 & 0.90 \\
45 & $2.89\times 10^4$ & $4.02\times 10^4$ & 1.99 & 1.01 \\
60 & $6.94\times 10^4$ & $8.93\times 10^4$ & 4.22 & 0.94 \\
\bottomrule
\end{tabular}
\end{table}

The measured exponents lie in the narrow range $0.90$--$1.01$, empirically supporting the predicted $O(P)$ per-sweep cost. Combined with Table~\ref{tab:speed}, this confirms that the resolution-dependent decay in speedup is driven by the growth in Schwarz iterations $N_{\mathrm{it}}$ rather than by super-linear per-sweep work. These results identify global Schwarz convergence, rather than local neural inference, as the main scalability bottleneck of the current implementation. This motivates multilevel coarse corrections, learned relaxation strategies, or accelerated Schwarz variants as natural directions for further improving large-scale performance. A further expected advantage of NEST is that, once the local neural operator has been trained, inference does not require evaluating or linearizing the constitutive law. Therefore, the cost of a NEST patch evaluation is largely independent of whether the underlying PDE contains a simple or complex constitutive model. This contrasts with nonlinear FEM solvers, whose cost can increase substantially with more complex material laws due to quadrature, stress evaluation, tangent assembly, and nonlinear iterations.

\section{NEST inference: pseudo-code and algorithmic details}
\label{app:pseudocode}
This appendix complements Section~\ref{subsec:schwarz} with a self-contained
algorithmic description of the NEST inference pipeline.
Algorithm~\ref{alg:nest-prep}, Phase 1 performs offline pre-processing of an arbitrary
voxelized domain $\Omega$, and Algorithm~\ref{alg:nest-schwarz}, Phase 2 performs the
Schwarz iteration itself. We adopt the notation of Section~\ref{sec:nest}:
the macro voxel grid has cells $\{C_k\}$ of edge length $h$ with $N_d$ cells
along axis $d \in \{1,2,3\}$, active node set
$\mathcal{V} \subseteq \mathbb{R}^3$, macro Dirichlet boundary
$\partial\Omega \subset \mathcal{V}$, overlapping
$3 \times 3 \times 3$ patches $\{\Omega_p\}_{p=1}^{P}$ with active nodes
$\mathcal{V}_p$, edges $\mathcal{E}_p$, and boundary partition
$\partial\Omega_p = \Gamma_p^{\mathrm{ext}} \cup \Gamma_p^{\mathrm{int}}$,
$\Gamma_p^{\mathrm{ext}} = \partial\Omega_p \cap \partial\Omega$. We write
$\{\chi_p\}_{p=1}^{P}$ for the partition of unity over $\mathcal{V}$,
$\mathcal{S}_\theta$ for the frozen local GNO of Section~\ref{sec:local},
$g : \partial\Omega \to \mathbb{R}^3$ for the macro Dirichlet datum, and
$\widetilde{g}$ for its inverse-distance$^{-2}$ extension to interior nodes
defined in Section~\ref{sec:local}. The per-axis partition-of-unity weight
is the piecewise-linear ramp
\begin{equation}
\tau(\xi;\, s, N) =
\begin{cases}
3\xi,        & 0 \le \xi < \tfrac{1}{3} \text{ and } s > 0,\\
3(1-\xi),    & \tfrac{2}{3} < \xi \le 1 \text{ and } s < N - 3,\\
1,           & \text{otherwise},
\end{cases}
\label{eq:tau}
\end{equation}
where $\xi \in [0,1]$ is the patch-reference coordinate along one axis, $s$
the patch anchor, and $N$ the number of macro cells on that axis.

\paragraph{Algorithmic clarifications.}

\textit{Partition-of-unity weights.} $\chi_p$ is the tensor product of three
per-axis ramps~\eqref{eq:tau}: linear from $0$ to $1$ on the first cell unless
flush-left ($s_d = 0$), constant $1$ on the central cell, and linear from $1$
to $0$ on the last cell unless flush-right ($s_d = N_d - 3$). Weights are
floored at $10^{-6}$ and renormalised at assembly so that
$\sum_p \chi_p(v) \equiv 1$ on every covered $v \in \mathcal{V}$.

\textit{Update style and damping.} The sweep is \textit{additive and Jacobi-style}:
in iteration $n+1$ all patches read the same $u^{(n)}$, with no Gauss--Seidel
sequential pass. The damping $\theta \in (0,1]$ relaxes the swept field
\textit{after} partition-of-unity assembly; $\theta = 1$ recovers undamped
additive Schwarz and is the default in all reported experiments.

\begin{algorithm}[htbp]
\caption{NEST Framework: Pre-processing and Schwarz Inference}
\label{alg:nest-main}

\textbf{Phase 1: Pre-processing of a voxelized domain $\Omega$}
\label{alg:nest-prep}
\vspace{1mm}
\begin{algorithmic}[1]
\Require macro voxel mask $S \in \{0,1\}^{N_1 \times N_2 \times N_3}$
         (largest face-connected component of the voxelizer output)
\Ensure  patches $\{\Omega_p\}_{p=1}^{P}$ with $\mathcal{V}_p$, $\mathcal{E}_p$,
         $\Gamma_p^{\mathrm{ext}}$, $\Gamma_p^{\mathrm{int}}$, $\chi_p$
\Statex
\State $\mathcal{V} \gets \{v : \exists\, C_k \in \mathrm{adj}(v) \text{ with } S_k = 1\}$
\State $\partial\Omega \gets \{v \in \mathcal{V} : \exists\, C_k \in \mathrm{adj}(v)
       \text{ with } S_k = 0 \text{ or } C_k \notin S\}$
\For{$(s_1, s_2, s_3)$ with $0 \le s_d \le N_d - 3$ for $d \in \{1,2,3\}$}
    \State $\Omega_p \gets$ cells of $S$ in the $3^3$ block anchored at $(s_1, s_2, s_3)$
    \State $\mathcal{V}_p,\, \partial\Omega_p \gets$ same 8-cell stencil restricted to $\Omega_p$
    \If{$\mathcal{V}_p = \emptyset$ \textbf{or} $\partial\Omega_p \cap \mathcal{V}_p = \emptyset$}
        \State \textbf{continue} 
    \EndIf
    \State $\Gamma_p^{\mathrm{ext}} \gets \partial\Omega_p \cap \partial\Omega$,\ \
           $\Gamma_p^{\mathrm{int}} \gets \partial\Omega_p \setminus \Gamma_p^{\mathrm{ext}}$
    \State $\mathcal{E}_p \gets \{(i,j) \in \mathcal{V}_p^2 :\ 0 < \|x_i - x_j\| \le 1.5\,h\}$
    \State $\chi_p(v) \gets \tau(\xi_1; s_1, N_1)\,\tau(\xi_2; s_2, N_2)\,\tau(\xi_3; s_3, N_3)$
           for each $v \in \mathcal{V}_p$ with patch-reference $(\xi_1, \xi_2, \xi_3) \in [0,1]^3$
\EndFor
\State \Return $\{\Omega_p\}_{p=1}^{P},\ \mathcal{V},\ \partial\Omega$
\end{algorithmic}

\vspace{3mm}
\hrule
\vspace{3mm}

\textbf{Phase 2: NEST Schwarz inference (additive, Jacobi-style)}
\label{alg:nest-schwarz}
\vspace{1mm}
\begin{algorithmic}[1]
\Require patches $\{\Omega_p\}_{p=1}^{P}$ from Phase 1;
         frozen $\mathcal{S}_\theta$; macro datum $g$;
         tolerance $\varepsilon$; damping $\theta \in (0,1]$;
         max iterations $N_{\max}$
\Ensure  global displacement $u^\star : \mathcal{V} \to \mathbb{R}^3$
\Statex
\State $u^{(0)} \gets \widetilde{g}$ on $\mathcal{V}$
       \Comment{$u^{(0)}\big|_{\partial\Omega} = g$}
\For{$n = 0, 1, \dots, N_{\max} - 1$}
    \State $a \gets 0 \in \mathbb{R}^{|\mathcal{V}| \times 3}$,\ \
           $w \gets 0 \in \mathbb{R}^{|\mathcal{V}|}$
    \For{$p = 1, \dots, P$}
        \State $g_p^{(n)}\big|_{\Gamma_p^{\mathrm{ext}}} \gets g$,\ \
               $g_p^{(n)}\big|_{\Gamma_p^{\mathrm{int}}} \gets u^{(n)}$
               \Comment{Eq.~\eqref{eq:schwarz_local}}
        \State $\hat u_p^{(n+1)} \gets
                \mathcal{S}_\theta\bigl(\mathcal{V}_p,\, \mathcal{E}_p,\, \widetilde{g_p^{(n)}}\bigr)$
        \State $a[\mathcal{V}_p] \mathrel{+}= \chi_p \odot \hat u_p^{(n+1)}$,\ \
               $w[\mathcal{V}_p] \mathrel{+}= \chi_p$
    \EndFor
    \State $u^{(n+\frac12)}(v) \gets a(v) / w(v)$ for $v \in \mathcal{V}$ with $w(v) > 0$
           \Comment{Eq.~\eqref{eq:schwarz_assembly}}
    \State $u^{(n+1)} \gets \theta\, u^{(n+\frac12)} + (1 - \theta)\, u^{(n)}$,\ \
           $u^{(n+1)}\big|_{\partial\Omega} \gets g$
    \State Compute $\Delta^{(n+1)}$ as in Eq.~\eqref{eq:delta};\
           \textbf{if} $\Delta^{(n+1)} < \varepsilon$ \textbf{then break}
\EndFor
\State $u^\star \gets u^{(n+1)}$
\State \textbf{Optional gradient pass:} for each $p$, feed $u^\star\big|_{\partial\Omega_p}$
       to the dedicated $\nabla u_{ij}$ surrogate and assemble per component through $\{\chi_p\}$.
\State \Return $u^\star$
\end{algorithmic}
\end{algorithm}

\end{document}